\documentclass[letterpaper, 10 pt, conference]{ieeeconf}  

\IEEEoverridecommandlockouts                              

\usepackage{graphics} 
\usepackage{amsmath} 
\usepackage{amssymb}  

\newtheorem{prop}{Proposition}
\newtheorem{defn}{Definition}

\newtheorem{rem}{Remark}
\usepackage{xcolor}
\usepackage{graphicx}
\usepackage[ruled,algo2e,linesnumbered]{algorithm2e}
\SetKwInput{KwInput}{Input}                
\SetKwInput{KwOutput}{Output}              

\title{\LARGE \bf
Robust Helicopter Ship Deck Landing With Guaranteed Timing Using Shrinking-Horizon Model Predictive Control
}

\author{Philipp Schitz$^{1,2}$ and Paolo Mercorelli$^{2}$ and Johann C. Dauer$^{1}$
\thanks{$^{1}$Institute of Flight Systems, German Aerospace Center (DLR), 38108 Braunschweig, Germany (e-mail: philipp.schitz@dlr.de; johann.dauer@dlr.de).}%
\thanks{$^{2}$Institute for Production Technology and Systems, Leuphana University of Lueneburg, 21335 Lueneburg, Germany (e-mail: paolo.mercorelli@leuphana.de).}%
}
\begin{document}

\maketitle
\thispagestyle{empty}
\pagestyle{empty}

\begin{abstract}

We present a runtime efficient algorithm for autonomous helicopter landings on moving ship decks based on Shrinking-Horizon Model Predictive Control (SHMPC). First, a suitable planning model capturing the relevant aspects of the full nonlinear helicopter dynamics is derived. Next, we use the SHMPC together with a touchdown controller stage to ensure a pre-specified maneuver time and an associated landing time window despite the presence of disturbances. A high disturbance rejection performance is achieved by designing an ancillary controller with disturbance feedback. Thus, given a target position and time, a safe landing with suitable terminal conditions is be guaranteed if the initial optimization problem is feasible. The efficacy of our approach is shown in simulation where all maneuvers achieve a high landing precision in strong winds while satisfying timing and operational constraints with maximum computation times in the millisecond range.
 
\end{abstract}

\section{Introduction}

Autonomous helicopters are an increasingly interesting platform for offshore logistics due to their high payload capacity and ability to hover. One crucial element of offshore missions is the ship deck landing which is notoriously challenging due to the ship movement and intense wind conditions. 
For the success of the landing, it is crucial to have timing and constraint satisfaction guarantees. In order to land in a specified time, it is necessary to compute a viable helicopter trajectory to the predicted ship deck position which must terminate at a suitable state for subsequent touchdown. Furthermore, on-board runtime efficiency is vital for flexibility across various landing scenarios. 

A promising approach for these requirements is Model Predictive Control (MPC), an optimization based strategy where an internal model is used to optimize control inputs over a given prediction horizon. However, standard receding-horizon MPC is not suitable for this task as the required terminal state for a landing will not be an equilibrium of the helicopter system. Particularly, the touchdown should happen with a downward velocity onto the ship, as this will allow for faster and more dynamic landings, extending the range of environmental conditions for landing.

This can be achieved with variable-horizon MPC strategies where the prediction horizon length varies across time steps. In \cite{perssonVariablePredictionHorizon2021}, a two-layer approach is used to land a UAV on a moving target. The lower layer is a Quadratic Program (QP) with a fixed prediction horizon while the upper layer optimizes over different horizon lengths. The approach by \cite{ngoVariableHorizonModel2022a} uses a mixed-integer QP formulation for a helicopter ship deck landing with linearized high-order models. However, both approaches have a high computational burden, thus hindering their use in real-time applications. A slightly different approach is taken in \cite{zhaoDifferentialFlatnessBasedApproachAutonomous2022} where the sample time of the discrete linear dynamics is optimized to vary the horizon length. While integer decision variables are avoided, this leads to a highly nonlinear formulation, producing similar problems for on-board use. However, when the maneuver time is specified beforehand, the variable-horizon problem reduces to a Shrinking-Horizon MPC (SHMPC), leading to a sequence of varying QPs which are much faster to solve. This approach was taken in \cite{greerShrinkingHorizonModel2020a}, where its efficacy was shown in extensive numerical experiments with representative dynamics and environmental conditions. Still, none of these approaches provide a sufficiently low computational burden for onboard use, robust timing guarantees or recursive feasibility. 

Our contribution is to apply a previously developed robust and runtime efficient shrinking-horizon strategy \cite{schitzRobustManeuverPlanning2024} to the autonomous ship deck landing problem. In particular, we use the robust finite time completion property of the algorithm together with a carefully designed terminal set to guarantee a given maneuver time, even in the presence of disturbances. Additionally, we propose a simplified control-oriented helicopter model for trajectory planning in strong winds and achieve a high disturbance rejection performance using a disturbance feedback. 

The rest of the paper is structured as follows: Section \ref{sec:method} describes the problem and reviews the robust SHMPC strategy. In Section \ref{sec:modeling}, the simplified model is derived which is afterwards used in Section \ref{sec:SHMPC_Design} for the design of the SHMPC components. Simulations are performed in Section \ref{sec:simulation} before concluding the paper in Section \ref{sec:conclusion}.

\subsection{Notation}

Given two sets $\mathcal{S}_1$, $\mathcal{S}_2$, set addition and erosion are defined as $\mathcal{S}_1 \oplus \mathcal{S}_2 = \{a + b \, | \, a \in \mathcal{S}_1, \, b \in \mathcal{S}_2 \}$ and $\mathcal{S}_1 \ominus \mathcal{S}_2 = \{a \, | \, a \oplus \mathcal{S}_2 \subseteq \mathcal{S}_1 \}$, respectively. 
We denote a block diagonal stacking of matrices $A$ by $\mathrm{diag}(A_1,\ldots,A_n)$. The $n$-dimensional identity matrix is denoted by $I_n$. We interpret $0_{n \times m}$ as a $n \times m$-dimensional matrix of zeros. 
The set of natural numbers ranging from $l$ to $u$ is written as $\mathbb{N}_l^u$.

\section{Problem Description and Preliminaries}\label{sec:method}

We consider the problem of landing an unmanned helicopter at a specific target region in a given time while satisfying linear constraints in the presence of additive disturbances. In conjunction with a ship motion estimator, this skill enables dynamic ship deck landings during high sea states. In this work, we assume that the landing target location and time are known, i.e., they were already chosen based on an existing ship motion prediction. 

\begin{figure}
	\begin{center}
		\includegraphics[width=0.9\linewidth]{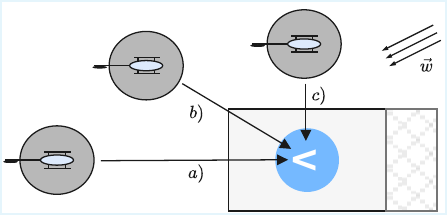}    
		\caption{Common ship deck landing approaches for  helicopters. Maneuver $a)$ is called straight-in, $b)$ a diagonal and $c)$ a lateral approach. The wind velocity vector is denoted by $w$.} 
		\label{fig:scenario}
	\end{center}
\end{figure}

We focus our attention on well-established \cite{HelicopterOperationsShips2017} ship deck landing maneuvers depicted in Figure \ref{fig:scenario}. Maneuver a) is a straight-in approach where the landing is performed from behind the ship without lateral displacement.  Maneuver c) starts from a hover or steady forward flight alongside the ship and then approaches the ship laterally. Lateral maneuvers are particularly useful in the presence of strong crosswinds as the turbulent airwake produced by the ships superstructure can be avoided while waiting for a quiescent period of the ship deck motion.  Maneuver b) is a combination of the previous two, resulting in a diagonal approach direction. For all of the mentioned landing approaches, the helicopters and ships headings are constant and aligned. 

In this work, we apply the SHMPC approach from \cite{schitzRobustManeuverPlanning2024} which comes with two advantages: First, it guarantees a specified maneuver time and operational constraints, even in the presence of additive disturbances, using a tube-based design. Second, the use of move blocking can achieve very large prediction horizons with low computational burden. This enables the consideration of practically relevant maneuver times without sacrificing theoretical guarantees. In the following, the approach is briefly outlined for completeness. 

\subsection{Robust Shrinking-Horizon MPC with Move Blocking}%

\begin{figure}
	\begin{center}
		\includegraphics[width=0.9\linewidth]{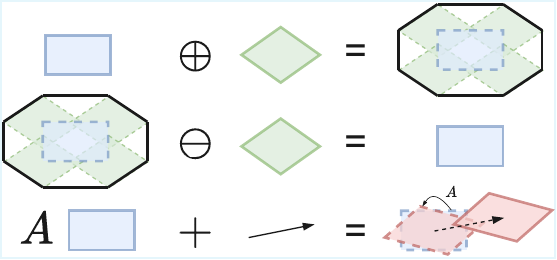}    
		\caption{Visual representation of the used set operations where dots mark the coordinate origin of the respective sets. The top row shows set addition $\mathcal{S}_1 \oplus \mathcal{S}_2$, the middle row set erosion $\mathcal{S}_1 \ominus \mathcal{S}_2$, and the bottom row an affine transformation $A \mathcal{S}_1 + b$.} 
		\label{fig:sets}
	\end{center}
\end{figure}

Consider the following linear discrete-time model where $k$ denotes a sample at time $k\tau$ with sample time $\tau$:
\begin{equation}\label{eq:system}
	x_{k+1} = A x_k + B u_k + W d_k
\end{equation}
subject to state-input and terminal constraints 
\begin{equation}\label{eq:sys_constraints}
	\forall k \in \mathbb{N}_0^{N_0-1}: \; [u_k, x_k]^T \in \mathcal{F},\; x_{N_0} \in \mathcal{X}_T,
\end{equation}
with state $x \in \mathbb{R}^n$, input $u \in \mathbb{R}^m$ and an unknown but bounded disturbance $d \in \mathcal{D} \subset \mathbb{R}^l$. The number of time steps within the maneuver is denoted as $N_0 \in \mathbb{N}^+$. To ensure robustness of the proposed landing algorithm, a tube-based MPC formulation is employed. This concept is based on establishing a nominal system with state $z$ and input $v$
\begin{equation}\label{eq:nominal_system}
	z_k = A z_k + B v_k
\end{equation}
without disturbances and an ancillary controller designed to keep the true system state $x_k$ within a set $\mathcal{Z}$ of the nominal trajectory $z_k$. In this work, $\mathcal{Z}$ is constant and given by a Robust Positive Invariant (RPI) set.
\begin{defn}[Robust Positive Invariance]
	Let $A_K = A-BK$ be such that $x_{k+1} = A_K x_k$ is stable. A set $\mathcal{Z}$ is RPI if $A_K \mathcal{Z} \oplus \mathcal{D} \subseteq \mathcal{Z}$.
\end{defn}
Based on this definition, we can state the following:
\begin{prop}[Proposition 1 in \cite{mayneRobustModelPredictive2005}] \label{prop:RPI}
	Let $\mathcal{Z}$ be a RPI set for system \eqref{eq:system}. If $x_0 \in z_0 \oplus \mathcal{Z}$ and 
	\begin{equation}\label{eq:MPC_control}
		u_k = v_k - K(x_k - z_k),
	\end{equation}
	then $x_{k} \in z_{k} \oplus \mathcal{Z}$ for all $d_k \in \mathcal{D}$ and $k \in \mathbb{N}$.
\end{prop}
Proposition 1 states that, using the controller gain $K$ from Definition 1, the control law \eqref{eq:MPC_control} shifts $\mathcal{Z}$ along the nominal trajectory $z_k$ to ensure that the trajectory error $x_k - z_k$ remains bounded for all disturbance realizations within $\mathcal{D}$.
This allows us to perform the MPC optimization over $z_k$ and $v_k$ in
\eqref{eq:nominal_system} but with tightened constraint sets
\begin{equation}\label{eq:tight_constraints}
	\bar{\mathcal{F}} = \mathcal{F} \ominus (\mathcal{Z} \times K\mathcal{Z}), \; \bar{\mathcal{X}}_T = \mathcal{X}_T \ominus \mathcal{Z}.
\end{equation}
Thus, if the nominal state $z$ and input $v$ remain within the tightened constraints $\bar{\mathcal{F}}$, the true state $x$ and input $u$ will not violate the original constraints $\mathcal{F}$. A visual representation of the set operations necessary for the RPI set computation and tightening procedure can be found in Figure \ref{fig:sets}. 

Next, we employ move blocking \cite{cagienardMoveBlockingStrategies2007} to extend the prediction horizon length without increasing the number of decision variables in the optimization problem or sacrificing theoretical guarantees.
The main idea of move blocking is to decouple the prediction horizon from the number of decision variables within the optimization by allowing an input to be held constant over specific time interval. This is formalized using a blocking matrix $M$. 
The following is an example of a blocking matrix reducing the input vector $U = [u_0,\ldots,u_{N-1}]^T$ over the prediction horizon of $N=4$ to $\bar{N}=2$ decision inputs denoted by $\bar{U} := [\bar{u}_0,\ldots,\bar{u}_{\bar{N}-1}]^T$:
\begin{equation*}
	\begin{bmatrix}
		u_0 \\
		u_1 \\
		u_2 \\
		u_3 \\
	\end{bmatrix}= (M \otimes I_m) \begin{bmatrix}
		\bar{u}_0 \\
		\bar{u}_1 \\
	\end{bmatrix} \text{ with } M=\begin{bmatrix}
		1 & 0 \\
		1 & 0 \\
		1 & 0 \\
		0 & 1 \\
	\end{bmatrix},
\end{equation*}
where the operator $\otimes$ denotes the Kronecker product. With the initial condition error defined as $\varepsilon_0 = x_k - z_0$, the SHMPC with move blocking can now be stated as:
\begin{equation} \label{eq:MB_SHMPC}
	\begin{aligned}		
		\;\min_{\bar{V},z_0} \; & \sum_{i=0}^{N_k-1} [z_i, v_i]^T H [z_i, v_i] + z_{N_k}^T P z_{N_k} + \varepsilon_0^T P_0 \varepsilon_0,\\
		\textrm{s.t.}  \; &  \varepsilon_0 \in \mathcal{Z}, \; z_{N_k} \in \bar{\mathcal{X}}_T, \; [z_i,v_i]^T \in \bar{\mathcal{F}}, \\
		& V = [v_0,\ldots,v_{N_k-1}]^T = (M_k \otimes I_m ) \bar{V}, \\
		& z_{i+1} = A z_i + B v_i, \; i \in \mathbb{N}_0^{N_k-1},
	\end{aligned}
\end{equation}
where $i$ denotes the prediction time step, $H$ is positive semi-definite, $P$ and $P_0$ are positive definite, $\bar{V} = [\bar{v}_0,\ldots,\bar{v}_{\bar{N}}]^T$ is the reduced input vector and the time-varying prediction horizon is defined as $N_k := N_0 - k$. With the optimal solution at time $k$ given by nominal input trajectory $V^*(k) = [v_0^*(k),\ldots,v_{N_k-1}^*(k)]^T$ and the initial reference state $z_0^*(k)$, the control law is $u_k = v_0^*(k) - K(x_k - z_0^*(k))$.
As is shown in \cite{schitzRobustManeuverPlanning2024}, applying this procedure guarantees that, if the initial problem is feasible, the terminal set in reached in exactly $N_0$ steps while remaining recursively feasible despite presence of disturbances. 

\section{Control-Oriented Helicopter Modeling}\label{sec:modeling}

We begin by introducing the general form of the helicopter dynamics with state $x_H = [p,v,\eta,\omega,\chi]^T$, where $p = [p_x, p_y, p_z]^T$, $v= [v_x, v_y, v_z]^T$, $\eta = [\phi,\theta,\psi]^T$, and $\omega \in \mathbb{R}^3$ denote the inertial position and velocity, the Euler angles and the body-fixed angular rates, respectively. The state $\chi$ represents any additional states that describe main or tail rotor dynamics such as blade flapping, inflow or engine behavior. The actual number of states $\chi$ varies with the models fidelity. The input $u_H = [u_\text{lon},u_\text{lat},u_\text{ped},u_\text{col}]^T$ consists of longitudinal and lateral cyclic, pedal, and collective inputs. The dynamics in their general form are thus given by
\begin{subequations}\label{eq:full_dynamics}
	\begin{align}
		\dot{p} &= v, \quad
		\dot{v} = m^{-1} \mathcal{R} {F}(x_H,u_H) + e_z g \label{subeq:transl_dynamics}\\
		\dot{\eta}& = \Psi \omega, \quad 
		\dot{\omega} = -(\omega \times J{\omega}) + {\tau}(x_H,u_H) \\
		\dot{\chi} &= f_\chi(x_H,u_H) 
	\end{align}
\end{subequations}
where $\Psi$ encodes the kinematic relation between the Euler angles and the angular rates, 
$\mathcal{R} \in SO(3)$ is the rotation matrix from body-fixed to local inertial coordinates, $e_z = [0, 0, 1]^T$, $g$ denotes the gravitational constant, and $F$ and $\tau$ denote general forces and torques, respectively. 

\subsection{Model simplification}

A key step to arrive at a runtime efficient trajectory planning formulation of the landing problem is the appropriate simplification of the model. In \cite{kooOutputTrackingControl1998}, a approximate flatness-based linearization of the helicopter dynamics is derived taking into account the translational and rotational states but no additional states $\chi$. 

In this work, we propose a similar linearization for the translational dynamics, the attitude dynamics however represent an existing inner attitude controller imposing a reference dynamics on $\eta$. This way, we defer the problem of stabilizing the complex helicopter dynamics to specialized design procedures that consider the influence of $\chi$. Importantly, this also allows for a full-state feedback in the MPC loop as the reference states are known, while some states within $\chi$ might not be measurable. Additionally, we are agnostic towards the actual controller used, improving the adaptability of the trajectory planner across multiple systems.

As the ship and helicopter yaw angles remain constant throughout the landing, we assume that the helicopter is stabilized at $\psi = 0$ without loss of generality.
We separate the translational subsystem \eqref{subeq:transl_dynamics} into a nominal part and a lumped disturbance $d(t) \in \mathcal{D}$ as follows: 
\begin{equation}\label{eq:simple_transl_dynamics}
	\begin{aligned}
		\dot{v} &= -\mathcal{R} e_z T + e_z g - D v + d(t) 
	\end{aligned}
\end{equation}
where $T := \alpha u_\text{col}$ denotes a nominal thrust acceleration which points in the body-fixed $z$-direction specified by $\mathcal{R} e_z$, $D = \text{diag}(D_x,D_y,D_z)$ contains linear drag coefficients, and $d(t)$ represents all remaining influences. By treating the thrust $T$ and attitude commands $\phi_c$, $\theta_c$ as the inputs to the translational subsystem, we can apply the input transform 
\begin{equation}\label{eq:input_transformation}
	a_c := -\mathcal{R} e_z T + e_z g
\end{equation}
to arrive at the linear representation $\dot{v} = a_c - D v + d(t)$.
We hence use $a_c = [a_{c,x}, a_{c,y}, a_{c,z}]^T$ during trajectory planning and recover the thrust and attitude commands using the inverse transform given by
\begin{equation}\label{eq:dynamics_transform}
	\begin{gathered}
		T = \sqrt{a_{c,x}^2+a_{c,y}^2+(a_{c,z}-g)^2}, \\
		\theta_c = \arctan\left(\frac{a_{c,x}}{a_{c,z}-g}\right), \quad
		\phi_c = \arcsin\left(\frac{a_{c,y}}{T}\right).
	\end{gathered}
\end{equation}
For the attitude subsystem, we assume the following second-order reference dynamics for $\phi$ and $\theta$:
\begin{equation}\label{eq:attititude_reference_dynamics}
	\begin{array}{lll}		
		\ddot{\phi}_r = -\omega_\phi^2 \dot{\phi}_r - 2 \omega_\phi \zeta_\phi (\phi_r - \phi_c), \\
		\ddot{\theta}_r = -\omega_\theta^2 \dot{\theta}_r - 2 \omega_\theta \zeta_\theta (\theta_r - \theta_c)
	\end{array}
\end{equation}
with reference states $\phi_r,\theta_r$, bandwidths $\omega_\phi, \omega_\theta$ and damping coefficients $\zeta_\phi, \zeta_\theta$. From \eqref{eq:dynamics_transform} it is apparent that $\theta_c$ is driven by $a_{c,x}$ and $\phi_c$ is driven by $a_{c,y}$. We therefore propose the following approximation for the reference dynamics of $\phi_r,\theta_r$ in terms of reference accelerations $a_{r,x}, a_{r,y}$ as
\begin{equation}\label{eq:acc_reference_dynamics}
	\begin{array}{lll}		
		\dot{j}_{r,x} = -\omega_\theta^2 j_{r,x} - 2 \omega_\theta \zeta_\theta (a_{r,x} - a_{c,x}), \\
		\dot{j}_{r,y} = -\omega_\phi^2 j_{r,y} - 2 \omega_\phi \zeta_\phi (a_{r,y} - a_{c,y}),
	\end{array}
\end{equation}
where $j_{r,x} := \dot{a}_{r,x}$, $j_{r,y} := \dot{a}_{r,y}$. 
Finally, we arrive at the linear control-oriented helicopter model with input $a_c$:
\begin{equation}\label{eq:control_model}
	\begin{aligned}
		&\dot{p} = v, \quad
		\dot{v} = [a_{r,x}, a_{r,y}, a_{c,z}]^T - Dv + d(t) \\
		&\dot{a}_{r,x} = j_{r,x}, \quad
		\dot{j}_{r,x} = -\omega_\theta^2 j_{r,x} - 2 \omega_\theta \zeta_\theta ({a}_{r,x} - a_{c,x}) \\
		&\dot{a}_{r,y} = j_{r,y}, \quad \dot{j}_{r,y} = -\omega_\phi^2 j_{r,y} - 2 \omega_\phi \zeta_\phi ({a}_{r,y} - a_{c,y})
	\end{aligned}
\end{equation} 
\begin{rem}
	The input $a_{c,z}$ acts directly on the vertical reference acceleration instead of passing through a second-order dynamics as $a_{c,x}$ and $a_{c,y}$. This allows a higher control bandwidth in the vertical dynamics while introducing approximation errors in the horizontal dynamics through \eqref{eq:acc_reference_dynamics}. This enables tighter bounds on vertical states and reduces the necessary landing time window at the cost of larger accuracy bounds on horizontal states.
\end{rem}

\subsection{Disturbance modeling}

During the derivation of the simplified model, we have made several approximations. The resulting errors are unified in the acceleration disturbance term $d(t)$ which is composed of model uncertainty introduced for the approximate feedback linearization in \eqref{eq:simple_transl_dynamics}, the tracking error $e_\eta = \eta - \eta_r$ of the inner controller w.r.t. the reference dynamics, and the approximation error from \eqref{eq:acc_reference_dynamics}. 


In general, the trajectory of $d(t)$ is unknown. However, experiments have shown that the disturbance can be separated into a constant term $\bar{d}$ and a time-varying part $d^v \in \mathcal{D}^v := \mathcal{D} - \bar{d}$:
\begin{equation}\label{eq:dist_decomp}
	d(t) = \bar{d} + d^v(t).
\end{equation}
Effects that can be attributed to $\bar{d}$ include accelerations acting due to nonzero trim conditions or influences from a slowly varying mean wind. Assuming that these effects are dominant in a hover, the influence of $\bar{d}$ is likely also significant when the helicopter is close to touchdown. Thus, if an estimate of $\bar{d}$ is available, it can be supplied to the planner in order to enhance the landing accuracy and expand the wind conditions in which the helicopter is able to land.

\subsection{State-space dynamics}

Augmenting model \eqref{eq:control_model} with the constant disturbance dynamics $\bar{d}_{k+1} = \bar{d}_{k}$, we arrive at a discrete-time state-space model with state $x = [x_\text{lon}, x_\text{lat}, x_\text{v}]^T$ and input $u = a_c$:
\begin{equation}\label{eq:ss_dynamics}
	\begin{gathered}
		x_{k+1} = A x_k + B u_k + W d^v_k
		, \\ A = \text{diag}(A_\text{lon},A_\text{lat},A_\text{v}), \quad B = \text{diag}(B_\text{lon}, B_\text{lat}, B_\text{v}), \\ W = \text{diag}(
		W_\text{lon}, W_\text{lat}, W_\text{v}),
	\end{gathered}
\end{equation}
with $x_i = [p_i,v_i,a_{r,i},j_{r,i},\bar{d}_i]$ for $i = \{\text{lon}, \text{lat}\}$, and $x_\text{v} = [p_z,v_z,\bar{d}_z]$. The discretized versions of the system matrix $A$, the input matrix $B$ and the disturbance matrix $W$ are derived from \eqref{eq:control_model}.

\section{SHMPC Design}\label{sec:SHMPC_Design}

In the section, we detail the design of the ancillary controller, the operational constraints and the terminal set to guarantee a safe landing in a specified time window.

\subsection{Disturbance-observer-based ancillary controller}

The role of the ancillary controller is to keep the actual system state $x$ close to its reference $z$ by counteracting the disturbances. In classical tube-based MPC, the ancillary controller is composed of a feedback and feedforward part as in \eqref{eq:MPC_control}. In order to improve the controllers performance, we propose an ancillary controller that also includes a disturbance estimate: 
\begin{equation}\label{eq:mpc_control}
	u_k = v_k - K (x_k - z_k) - K_d \hat{d}^v_k,
\end{equation}
where $K$ is the feedback gain and $K_d$ is the disturbance gain. Note that only the estimated varying part $\hat{d}^v_k = \hat{d}_k - \bar{d}$ according to the decomposition \eqref{eq:dist_decomp} is compensated as the constant part $\bar{d}$ is already accounted for by the nominal input $v_k$. The error dynamics with $\tilde{x}_k = x_{k} - z_{k}$ is then given by
\begin{equation*}
	\begin{array}{ll}
		\tilde{x}_{k+1} = A x_k + B u_k + W d^v_k - A z_k - B v_{k}
	\end{array}	
\end{equation*}
and by substituting \eqref{eq:mpc_control} we obtain
\begin{equation}
	\begin{array}{ll}
		\tilde{x}_{k+1} = (A - BK) \tilde{x}_k + W d^v_k - BK_d \hat{d}^v_k.
	\end{array}	
\end{equation}
In this work, we employ a discretized linear version of the disturbance estimator from \cite{wen-huachenNonlinearDisturbanceObserver2000}
\begin{subequations} 
\begin{align}
	s_{k+1} = & A_W s_k -L ((A-I_n) x_k + B u_k + W Lx_k) \\
	\hat{d}^v_k = & s_k + Lx_k
\end{align} \label{eq:disturbance_estimator}
\end{subequations}
with an auxiliary variable $s \in \mathbb{R}^{l}$ and the observer gain $L \in \mathbb{R}^{l \times n}$ chosen such that $A_W = I_l-LW$ is stable. With the estimation error defined as $\tilde{d}_{k} = d^v_k - \hat{d}^v_k$, we obtain a simple first-order estimation error dynamics that is driven by the disturbance change rate $\Delta d^v_k := d^v_{k+1} - d^v_k$:
\begin{align}\label{eq:disturbance_error_dynamics}
	\tilde{d}_{k+1} = A_W \tilde{d}_{k} + \Delta d^v_k.
\end{align}

\subsection{RPI set design}

Previous works incorporating a disturbance observer into the MPC formulation \cite{yanSurvivingDisturbancesPredictive2023a}, \cite{sunDisturbanceEstimationandExploitationBased2024} use the estimation error dynamics directly. A bound on $|\Delta d_k^v|$ is assumed to derive a maximum estimation error which then acts as the new unknown input. The bound on $|\Delta d_k^v|$ however may not exist in practice or might be large, leading to conservative estimates of the controller performance. To alleviate this restriction, we propose to augment the state vector to directly incorporate the  disturbance estimator dynamics. By rewriting \eqref{eq:disturbance_error_dynamics} as 
\begin{align}\label{eq:estimator_dynamics}
	\hat{d}_{k+1}^v = A_W \hat{d}_k^v + L W d_k^v,
\end{align}
we can see that the disturbance estimate is a low-pass filtered version of the true disturbance $d_k^v$. Therefore, if $d_k^v$ is bounded, so is $\hat{d}_{k}^v$. Introducing a new augmented state $\xi = [\tilde{x}^T, (\hat{d}^v)^T]^T$, we can bring the augmented error dynamics into the form required by Definition 1
\begin{equation}\label{eq:augmented_sys}
	\begin{gathered}
			\xi_{k+1} = \tilde{A} \xi_k + \tilde{B} d_k^v, \\
		  \tilde{A} = \begin{bmatrix}
				A - BK & - B K_d \\ 
				0_{l \times n} & A_W
			\end{bmatrix}, \quad \tilde{B} = \begin{bmatrix}
			W \\ 
			LW
			\end{bmatrix},
	\end{gathered}
\end{equation}
where $\tilde{A}$ and $\tilde{B}\mathcal{D}^v$ correspond to $A_K$ and $\mathcal{D}$, respectively. 
\begin{rem}
	Note that $K_d$ does not alter the eigenvalues of $\tilde{A}$ and hence the existance of a bounded RPI set is guaranteed as long as $K$ and $L$ are chosen such that $A-BK$ and $A_W$ are stable. However, the choice of $K_d$ significantly influences the size of the RPI set.
\end{rem}
For the computation of the RPI set $\mathcal{Z}_\xi$ for \eqref{eq:augmented_sys}, we use the well-established method from \cite{rakovicInvariantApproximationsMinimal2005}. Note that, by design, the error for $\bar{d}$ is always zero, so it can be ignored during the computation of the RPI set. To obtain the RPI set $\mathcal{Z}$ for system \eqref{eq:ss_dynamics}, we project $\mathcal{Z}_\xi$ back onto $\tilde{x}$.

\subsection{Constraints}

\begin{figure}
	\begin{center}
		\includegraphics[width=1.0\linewidth]{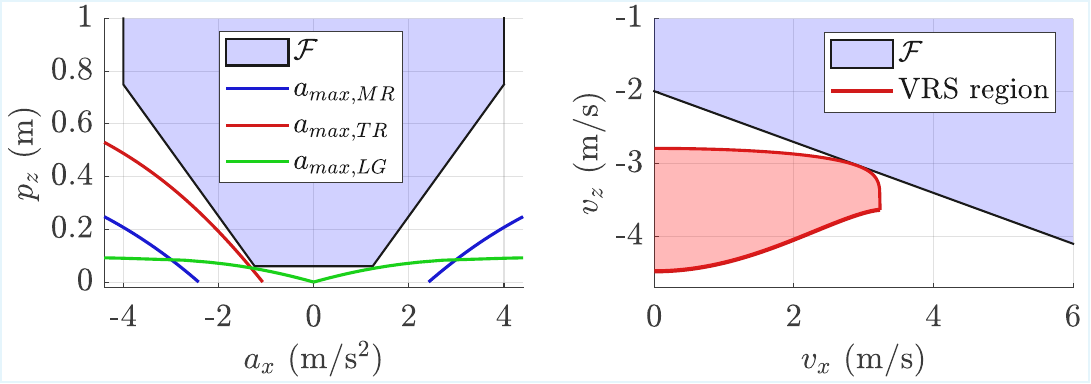}    
		\caption{Constraint set $\mathcal{F}$. Left: Attitude constraint and maximum accelerations for which geometric features intersect the ground. Right: Linear VRS constraint and exemplary nonlinear boundary computed based on \cite{johnsonVortex2005}.} 
		\label{fig:attitude_constraint}
	\end{center}
\end{figure}

For the landing maneuver, we first impose box constraints on the states and inputs as 
\begin{equation}\label{eq:box_constraints}
	\bar{x}_{lb} \leq \bar{x} \leq \bar{x}_{ub}, \quad \bar{u}_{lb} \leq \bar{u} \leq \bar{u}_{ub}
\end{equation}
where $\bar{x}_{lb}, \bar{x}_{ub}$ and $\bar{u}_{lb}, \bar{u}_{ub}$ denote lower and upper bounds on states and inputs, respectively. 
Additionally, we impose constraints to avoid vortex ring state (VRS) \cite{johnsonVortex2005} and height-dependent acceleration constraints to avoid a premature touchdown. Both types of constraints are linearly encoded:
\begin{subequations}\label{eq:vrs_att_contraints}
	\begin{gather}
		\gamma_{a,x} p_z + a_{r,x} \leq b_x, \; \gamma_{a,y} p_z + a_{r,y} \leq b_y, \\ \gamma_{vrs} v_x - v_z \leq b_{v,z},
	\end{gather}
\end{subequations}
where $\gamma$ and $b$ denote design parameters for slope and offset, respectively.
The VRS boundary constraint is roughly estimated from previous flight tests and restricts the descent velocity for low forward velocities (right plot of Figure \ref{fig:attitude_constraint}). The height-acceleration constraint is derived from the helicopters geometric features that limit its attitude close to the ground, i.e., the tail rotor (TR), main rotor (MR) or the landing gear (LG) (left plot of Figure \ref{fig:attitude_constraint}). 

\subsection{Terminal set}

The helicopter must not touch the ground before it admits a safe touchdown state. Using the height-attitude constraints, we ensured that touchdown cannot occur before the last moment of the MPC trajectory. Therefore, we introduce a touchdown stage that activates after the MPC trajectory ends. During this stage, a touchdown reference controller guides the nominal state to the ground while ensuring the helicopter remains in a safe state for touchdown. The ancillary controller is still active for this stage and thus the RPI set $\mathcal{Z}$ remains valid. The terminal set for the MPC is then the set of all initial nominal states that guarantee a safe touchdown using the touchdown reference controller. 

Let $z_0^{TD}$ denote the initial nominal state of the touchdown reference trajectory which coincides with the last nominal state of the MPC trajectory $z_{N_0}$. Further, let the touchdown reference controller be defined as 
\begin{equation}\label{eq:td_controller}
	v_k^{TD} = -K_{TD} (z_k^{TD} - z_{r}^{TD}),
\end{equation}
where $K_{TD}$ is a gain matrix and $z_{r}^{TD}$ is the touchdown controller setpoint. Substituting \eqref{eq:td_controller} into the nominal dynamics, we obtain
\begin{equation}\label{eq:touchdown_reference}
	\begin{gathered}
		z^{TD}_{k+1} = A_{TD} z_k^{TD} + B K_{TD} z_r^{TD}, \quad z_0^{TD} = z_{N_0},
	\end{gathered}
\end{equation}
with $A_{TD} := A-BK_{TD}$. In order to compute the terminal set, the necessary time to touchdown during this stage needs to be known. However, we cannot know where the true state lies within the RPI set a-priori and thus cannot determine the exact time to touchdown. We thus pick a maximum number of steps $N_{TD}$ which serves as a landing time window. During this window, touchdown could occur at any time, so the helicopter must remain in a safe touchdown state throughout. The set of safe touchdown states is encoded by $\mathcal{F}_{TD} \subset \mathbb{R}^n$. 
\begin{rem}\label{rem:TD_params}
	The choice of $K_{TD}$, $z_{r}^{TD}$, $N_{TD}$ and $\mathcal{F}_{TD}$ determines the aggressiveness, necessary landing time window, and size of the terminal set. For example, the smaller the landing time window, the smaller the terminal set will be. Similarly, the more aggressively the helicopter is allowed to touch down, the smaller the necessary landing time window.
\end{rem}

\begin{figure}
	\begin{center}
		\includegraphics[width=1.0\linewidth]{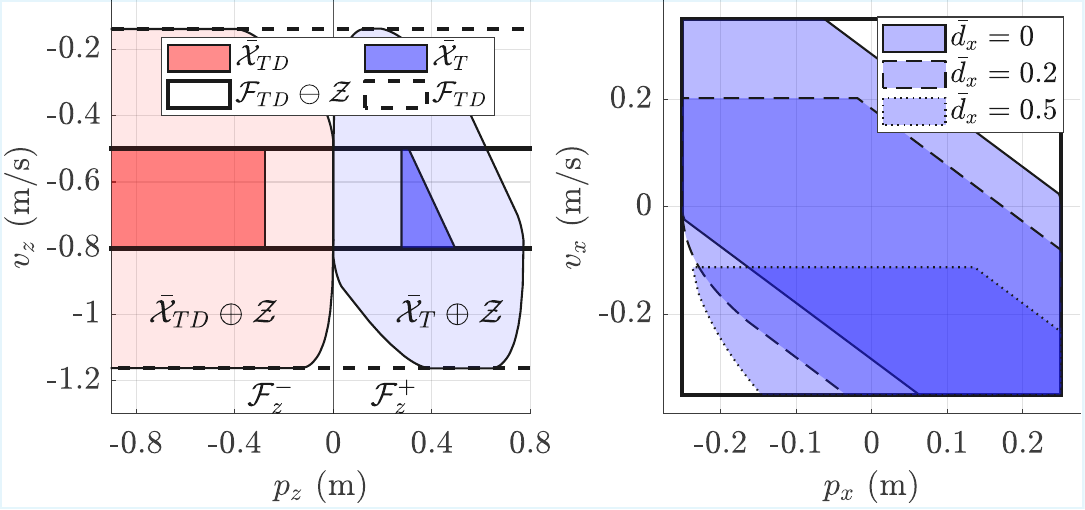}    
		\caption{Terminal sets for $N_{TD} = 50$. Left: Projections of  touchdown constraint set $\mathcal{F}_{TD}$, the starting set $\bar{\mathcal{X}}_{TD}$ encoding a successful touchdown, and the resulting terminal set $\bar{\mathcal{X}}_T$. Right: Projections of $\bar{\mathcal{X}}_T$ for different slices in $\bar{d}_x$.} 
		\label{fig:terminal_set}
	\end{center}
\end{figure}

We can now compute the set of all initial states $z_0^{TD}$ which satisfy the touchdown constraints $\mathcal{F}_{TD}$ for $N_{TD}$ steps using backward reachability analysis. First, we choose a starting set $\bar{\mathcal{X}}_{TD} \subset \mathbb{R}^{n}$ representing all nominal states for which the true system is guaranteed to have landed safely (see Figure \ref{fig:terminal_set}, left). Since, according to Proposition \ref{prop:RPI}, the true system is guaranteed to lie within the RPI set $\mathcal{Z}$ around the reference state, we need to ensure
\begin{equation}
	\bar{\mathcal{X}}_{TD} \oplus \mathcal{Z} \subseteq \mathcal{F}_{TD} \cap \mathcal{F}^-_{z},
\end{equation}
where $\mathcal{F}^-_{z} = \{ z^{TD} \in \mathbb{R}^{n} \, | \, p_z \leq 0 \}$. Setting $R_0 = \bar{\mathcal{X}}_{TD}$ and following the recursion 
\begin{equation}\label{eq:Terminal_set_recursion}
	\begin{aligned}
		R_{k+1} &= A_{TD}^{-1} (R_k - B K_{TD} z_{r,TD}) \cap \left(\mathcal{F}_{TD}  \ominus \mathcal{Z}\right)
	\end{aligned}	
\end{equation}
for $N_{TD}$ steps, we obtain the tightened terminal set as
\begin{equation}\label{eq:terminal_set}
	\bar{\mathcal{X}}_{T} = R_{N_{TD}} \cap \left(\mathcal{F}^+_{z} \ominus \mathcal{Z}\right)
\end{equation}
with the above-ground set $\mathcal{F}^+_{z} = \{ z^{TD} \in \mathbb{R}^{n} \, | \, p_z \geq 0 \}$. Therefore, if the nominal MPC trajectory terminates in $\bar{\mathcal{X}}_{T}$, we are guaranteed to reach $\bar{\mathcal{X}}_{TD}$ within $N_{TD}$ steps while satisfying $\mathcal{F}_{TD}$. Note that, as seen on the right plot of Figure \ref{fig:terminal_set}, the influence constant disturbance term $\bar{d}$ is also reflected in the terminal set. 
\begin{rem}
	This procedure also allows an evaluation of the chosen parameters from Remark \ref{rem:TD_params}. If $\bar{\mathcal{X}}_{T}$ is empty, then there exists no terminal set for which a safe landing within the time window can be guaranteed with the current choice of parameters, and thus different values must be chosen.
\end{rem}

\subsection{Landing Algorithm}

Algorithm \ref{alg1} summarizes the complete landing procedure. For the first $N_0$ steps, the SHMPC \eqref{eq:MB_SHMPC} is solved using $\bar{d}$, constraints \eqref{eq:box_constraints}, \eqref{eq:vrs_att_contraints}, and the terminal set computed from \eqref{eq:terminal_set} in line 3. After the MPC trajectory is finished, for a maximum of $N_{TD}$ steps, the touchdown reference dynamics \eqref{eq:touchdown_reference} is propagated using the input \eqref{eq:td_controller} in line 5. In both stages, the control input is updated according to \eqref{eq:mpc_control} and \eqref{eq:disturbance_estimator} in line 7.
\IncMargin{1em}
\begin{algorithm2e}
	\label{alg1}
	\caption{Timed Landing SHMPC}\label{alg:two}
	\For{$k=1$ \KwTo $N_0+N_{TD}$}{
		\uIf{$k \leq N_0$}{
		$(z_{k}, v_{k}) \gets$ \texttt{solveSHMPC}$(x_k,k)$\;	
		}
		\Else{
		$(z_k, v_{k}) \gets$ \texttt{propagateTD}$(z_{k-1})$\;
		}
		$u_{k} \gets$ \texttt{updateCtrl}$(x_{k}, z_k, v_{k})$\; 
	}
\end{algorithm2e}
\DecMargin{1em}

\section{Simulation}\label{sec:simulation}

For simulation, we use a nonlinear model of the form \eqref{eq:full_dynamics} of DLR's small-scale demonstrator midiARTIS where $\chi$ contains the flapping dynamics and an engine state. The parameters are extracted and interpolated from three linear models that were identified from experimental data at hover, 10~m/s and 20~m/s forward velocity. The inner loop controller is designed to follow the reference dynamics \eqref{eq:attititude_reference_dynamics}. For further information on both the system identification and the inner loop controller, see \cite{petitSystemIdentification2025}. 

We consider a landing on a ship moving at 5~m/s from different starting positions with a mean headwind of 8~m/s blowing from 30${}^\circ$.
Maneuver $a)$ (straight-in) begins at $[-30,0,30]$~m, $b)$ (diagonal) at $[-15,-10,25]$~m, and $c)$ (lateral) at $[0,-15,20]$~m relative to the ship. All maneuvers in this study are generated with $N_0 =$~495 prediction steps, $\bar{N} = 17$ blocked inputs and a sample time of 20~ms, resulting in a maneuver time of 9.9~s. The touchdown time window at the end of the trajectory was chosen to be 50 steps (the terminal set is shown in Figure \ref{fig:terminal_set}). We assume that the position is measured relative to the landing target. Thus, the position states of the terminal set are centered at the origin while velocities are shifted to match the ships velocity at touchdown. The matrices $K$ and $K_d$ are chosen as follows:
\begin{gather*}
	K = \text{diag}(K_\text{lon}, K_\text{lat}, K_\text{v}), \; \; 	K_d = \text{diag}(K_{d,\text{lon}}, K_{d,\text{lat}},  K_{d,\text{v}}), \\
	K_\text{lon} =  \left[\begin{IEEEeqnarraybox*}[][c]{,c/c/c/c/c,}
		0.6&1.4&K_a&0&K_a
	\end{IEEEeqnarraybox*}\right], \;
	K_\text{lat} = \left[\begin{IEEEeqnarraybox*}[][c]{,c/c/c/c/c,}
		0.4&1&K_a&0&K_a
	\end{IEEEeqnarraybox*}\right] \\
	K_\text{v} = \left[\begin{IEEEeqnarraybox*}[][c]{,c/c/c,}
		2&3&0
	\end{IEEEeqnarraybox*}\right], \; K_{d,\text{lon}} = K_{d,\text{lat}} = 1 + K_a, \; K_{d,\text{v}} = 1,
\end{gather*}
where $K_a = 0.5$ is an acceleration feedback gain. The MPC weighting matrices are tuned such that the reference trajectory remains as close as possible to the true state while ensuring smooth acceleration commands. Additionally, a high weight is placed on the terminal cost matrix $P$ to enhance landing precision. The touchdown reference controller is designed to level out the attitude of the helicopter while maintaining a descent velocity of 0.7 m/s. The position and time of the landing are chosen such that touchdown occurs during a quiescent period.

\begin{figure}
	\centering
	\includegraphics[width=0.9\linewidth]{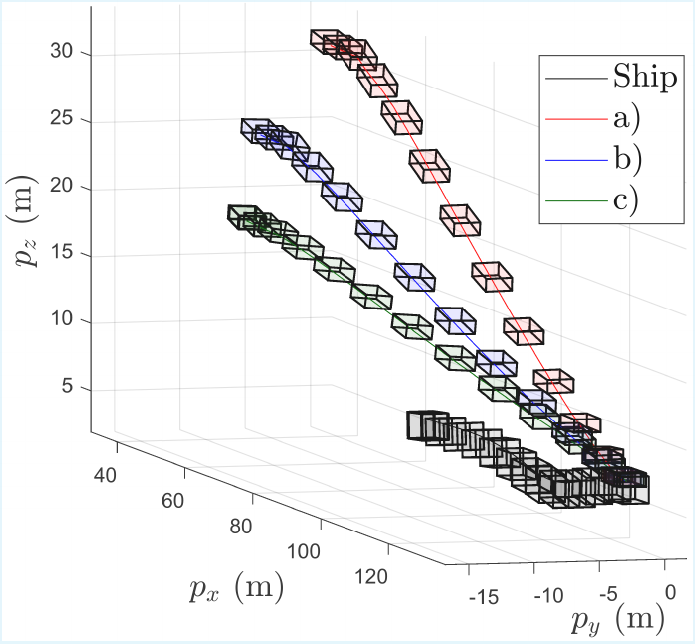}
	\caption{Ship and helicopter trajectories for the considered maneuvers.}
	\label{fig:traj}
\end{figure}

\begin{figure}
	\centering
	\includegraphics[width=1.0\linewidth]{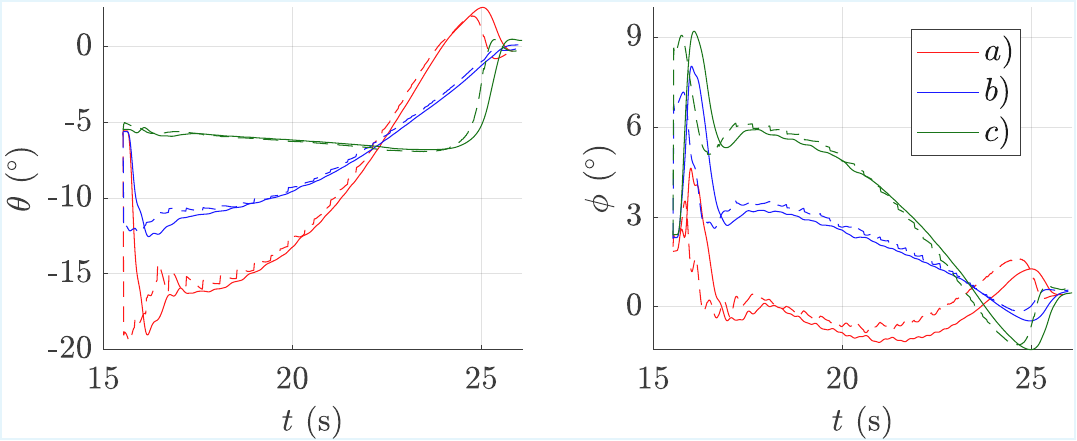}
	\caption{Pitch angles $\theta$ (left) and roll angles $\phi$ (right) trajectories during the landing maneuvers with true and commanded values to the attitude controller marked by solid and dashed lines, respectively. }
	\label{fig:attitude_ts}
\end{figure}

\begin{figure}
	\centering
	\includegraphics[width=1.0\linewidth]{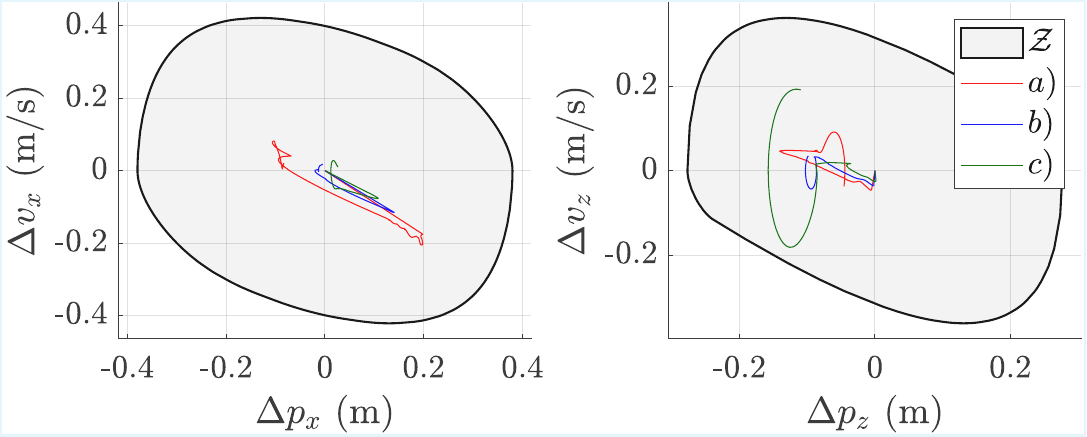}	
	\caption{Error trajectories in the longitudinal (left) and vertical (right) axis.}
	\label{fig:disturbance}
\end{figure}

%
Figure \ref{fig:traj} shows the 3D trajectories of all three maneuvers with attitude snapshots together with the ship motion while in Figure \ref{fig:attitude_ts} the commanded and true attitude values are plotted over time. Table \ref{tb:terminal_stats} reports ship-relative states in the moment of touchdown. The one-second time window after the 9.9~s maneuver was achieved without prematurely touching the ground, thus guaranteeing that the helicopter is admits a safe landing state before touchdown. This is also reflected is the terminal states in Table \ref{tb:terminal_stats} which all lie within their design bounds. Furthermore, as highlighted in Figure \ref{fig:disturbance}, the disturbances remain within their assumed bounds for the entire maneuver, ensuring the robustness and recursive feasibility guarantees of the SHMPC. Note that the original disturbance bound could not have been satisfied without the consideration of the constant disturbance part $\bar{d}$ in the optimization problem. Remarkably, it is $\bar{d}$ that enables the use of this method in the presence of strong winds.
This is also reflected in the bottom plots of Figure \ref{fig:disturbance} where the trajectory deviations never exceed the bounds of the RPI set $\mathcal{Z}$, underpinning the theoretical soundness of the method.
Thanks to the move blocking and constraint reduction approach in \cite{schitzRobustManeuverPlanning2024}, the maximum solver time across all simulations was 4~ms, which is well below the 20~ms sample time. The optimization was performed using the C++ interface of qpOASES \cite{ferreauQpOASESParametricActiveset2014} on a laptop with an Intel Core i7-11850H processor.
\begin{table}[hb]
	\begin{center}
		\caption{Ship-relative attitude $\Delta\phi$, $\Delta\theta$, horizontal position and velocity $\Delta p_{h}$ and $\Delta v_{h}$, and vertical velocity $\Delta v_{z}$ at touchdown, as well as maneuver time $T$.}\label{tb:terminal_stats}
		\begin{tabular}{c|c|c|c|c|c|c}
			Man. &  $|\Delta \phi|$ & $| \Delta \theta|$ &  $|\Delta p_{h}|$ & $|\Delta v_{h}|$ & $|\Delta v_{z}|$ & $T$ \\\hline
			$a)$ & $0.6^\circ$ & $0.2^\circ$ & $0.06$~m & $0.04$~m/s & $0.95$~m/s & 10.30~s \\
			$b)$ & $0.7^\circ$ & $0.6^\circ$ & $0.15$~m  & $0.33$~m/s & $0.91$~m/s & 10.44~s\\
			$c)$ & $0.6^\circ$ & $1.1^\circ$ & $0.14$~m  & $0.54$~m/s & $0.98$~m/s & 10.54~s\\
			\hline
			\textbf{max} & $4^\circ$ & $6^\circ$ & $0.60$~m& $ 0.60$~m/s& 1.15~m/s & 10.90~s\\
		\end{tabular}
	\end{center}
\end{table}%
\begin{figure}
	\centering
	\includegraphics[width=1.0\linewidth]{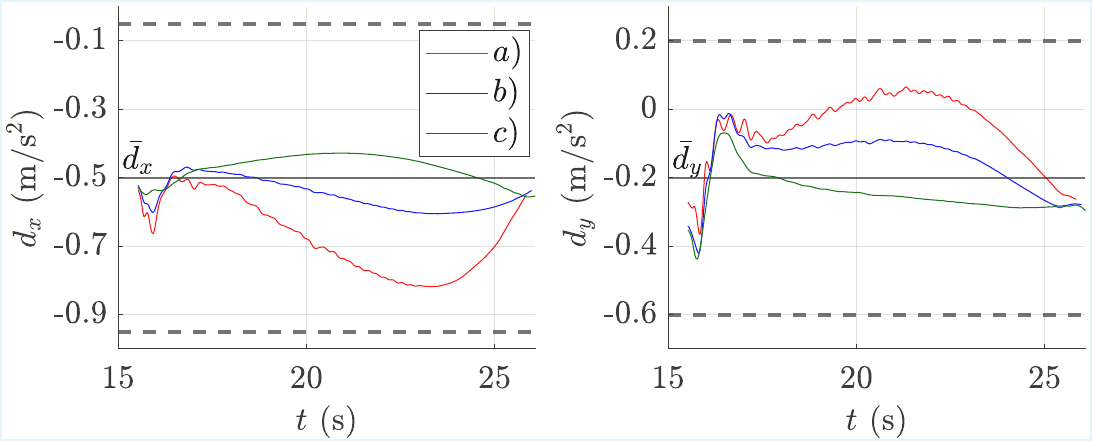}
	\caption{Longitudinal (left) and lateral (right) disturbance trajectories during the landing maneuvers. The boundaries of disturbance set $\mathcal{D} = \bar{d} + \mathcal{D}^v$ are marked by dashed lines.}
	\label{fig:ts_attitude}
\end{figure}%

\section{Conclusions and future work}\label{sec:conclusion}

In this paper, we proposed a helicopter ship deck landing procedure with landing time guarantees. We designed an appropriate planning model that captures the relevant translational dynamics while the attitude dynamics were approximated using a closed-loop reference dynamics. A key component is the use of Shrinking-Horizon Model Predictive Control with move blocking, resulting in runtime efficient, robust, and recursively feasible trajectory generation. Through a carefully designed terminal set, we were able to formally guarantee a time window for the touchdown on the ship, facilitating a safe landing. 
In future work, we plan to consider the prediction errors of the ship motion estimator on which the target position and necessary maneuver time are based. Furthermore, we intend to perform flight test trails to assess the performance of the method in practice.

\bibliography{mpc_landing.bib}             


\end{document}